\documentclass{article}
\usepackage{spconf,amsmath,graphicx,multirow,adjustbox}
\usepackage[rgb]{xcolor}
\usepackage{flushend}
\usepackage[font=small,skip=3pt]{caption}
\def\etal{\emph{et al}}

\title{Unsupervised Image Fusion Using Deep Image Priors}
\name{Xudong Ma, Paul Hill, Nantheera Anantrasirichai, Alin Achim}
\address{Visual Information Laboratory, University of Bristol, Bristol, UK}
\begin{document}
%\ninept
%
\maketitle
\begin{abstract}
A significant number of researchers have applied deep learning methods to image fusion. However, most works require a large amount of training data or depend on pre-trained models or frameworks to capture features from source images. This is inevitably hampered by a shortage of training data or a mismatch between the framework and the actual problem. Deep Image Prior (DIP) has been introduced to exploit convolutional neural networks' ability to synthesize the `prior' in the input image. However, the original design of DIP is hard to be generalized to multi-image processing problems, particularly for image fusion.
Therefore, we propose a new image fusion technique that extends DIP to fusion tasks formulated as inverse problems. Additionally, we apply a multi-channel approach to enhance DIP's effect further. 
The evaluation is conducted with several commonly used image fusion assessment metrics. The results are compared with state-of-the-art image fusion methods. Our method outperforms these techniques for a range of metrics. In particular, it is shown to provide the best objective results for most metrics when applied to medical images.
\end{abstract}
\section{Introduction}
\label{sec:intro}

Image fusion is the process of combining information from multiple images into a single representation \cite{2003Image}. It is widely used in many application domains, such as medical imaging \cite{2014Medical}, photography \cite{2004Multi}, and surveillance \cite{2018Multi}. A large amount of research has been done in this area. Conventionally, image fusion can be performed at pixel-level, feature-level, or decision-level \cite{2017pixel}. 
Pixel-level image fusion has been extensively studied because of its high efficiency \cite{2017pixel}. It can be performed using either fixed transform-based methods or via learnt Sparse Representations (SR)~\cite{2017pixel}. 
If using fixed transforms, the original images need to be decomposed into other domains using methods such as the Discrete Wavelet Transform (DWT) or the contourlet transform~\cite{2008Image}. Decomposition coefficients are fused using fusion-rules such as weighted-averaging \cite{2012Contrast} and maximum selection \cite{2012The}. 
Finally, the fused image is generated by taking the inverse transform of the fused coefficients.  
For SR-based methods, a dictionary is learnt before fusion. Researchers have proposed many ways to obtain the optimal dictionary. Most utilize either mathematical models or sample learning \cite{2010Dictionaries}. The input images are first decomposed and sparse coded during fusion. Different strategies have been developed to achieve this, for example, group SR \cite{2012Group} and gradient constrained SR \cite{2014Image}. Finally, the computed sparse coefficients are fused and reconstructed into a single image according to various fusion rules \cite{2010Joint}.

The fusion task can be seen as an inverse problem. Solving with mathematical models, various loss functions and optimization strategies have been proposed to get the best fusion results. For this kind of method, the loss function plays a vital role in the performance achieved. The $L_1$ norm is widely used as a regularization component of the loss function \cite{2002Mathematical}. However, it has some inherent weakness that often causes the final result to be over-smoothed. To address this, a non-convex penalty regularization technique was proposed in \cite{2020Image} and achieved state-of-the-art image fusion performance.
With the development of deep Convolutional Neural Networks (CNNs), researchers have started using such techniques for image fusion. Liu \etal~\cite{2017Multi} proposed the application of blurred image patches to train a decision map using CNNs. The fused image was generated based on the decision map and the input images. Li \etal~\cite{2018Infrared} enhanced Liu's method by introducing a multi-layer feature extraction technique. 
However, like most other deep learning based fusion models, they all depend on training data or pre-training. Nevertheless, training data is hard to obtain for image fusion tasks. Furthermore, it is also challenging to find well-fitted pre-trained models for some specific applications, such as the fusion of medical images. Most models are pre-trained on natural images, which are very different from medical images. This inherently limits performance in particular applications.

Unsupervised learning is a way to address the issues of current deep learning based fusion methods. This paper proposes a new technique, based on Deep Image Prior (DIP) \cite{2017Deep}, to fuse images in an unsupervised way. DIP assumes that a significant number of underlying image statistical priors can be obtained by the structure of CNNs even without any training \cite{2017Deep}. The authors employed an encoder-decoder network to verify their assumption on several single image restoration tasks. Although they also provided an example of applying DIP to multi-image processing problems, their design is still hard to be extended to problems such as image fusion. 

Here, we pose image fusion as an inverse problem and design a loss computed from the output and all source images directly. This makes it easy to tackle multi-image processing problems. In addition, this paper also proposes a multi-channel training strategy to enhance the image prior extraction ability of DIP. Finally, the performance of our approach is compared with the non-convex penalty based fusion method in~\cite{2020Image} and two deep learning based methods \cite{2018Infrared} \cite{2018DenseFuse}, which are recently published state-of-the-art model-based and data-driven image fusion techniques, respectively. Better results are obtained visually and objectively for the fusion of both mono-modal images and cross-modal images.
\section{Methodology}
\label{sec:methods}
\subsection{Inverse Problem Formulation}
\begin{figure}[htb]
  \centering
    \begin{tabular}{c@{\hspace{0.5pt}}c@{\hspace{0.5pt}}c@{\hspace{0.5pt}}c}
        % xudong guess: Adding vertical space after showing the images before //
        % \vspace{0.5pt} 
        \includegraphics[width=0.24\linewidth]{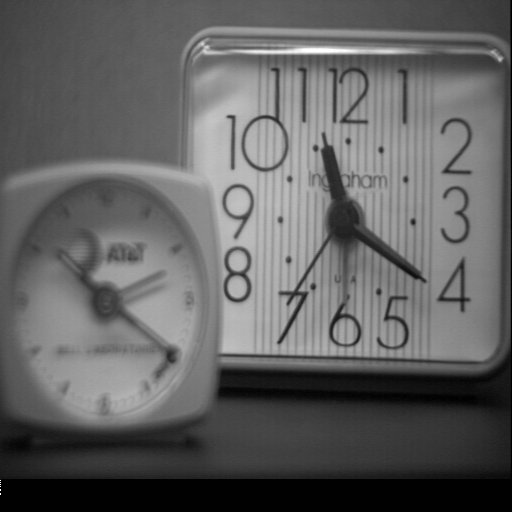}&
        \includegraphics[width=0.24\linewidth]{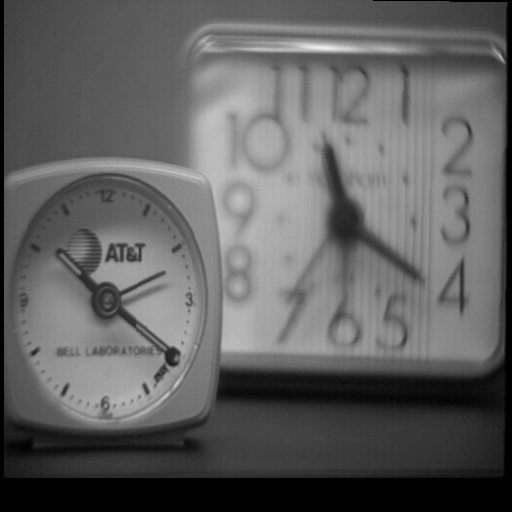}&
        \includegraphics[width=0.24\linewidth]{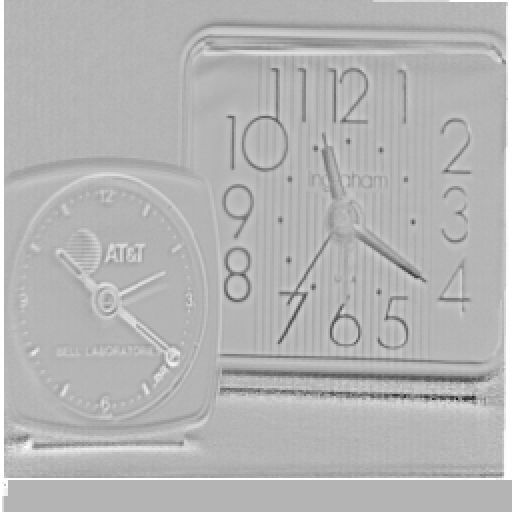}&
        \includegraphics[width=0.24\linewidth]{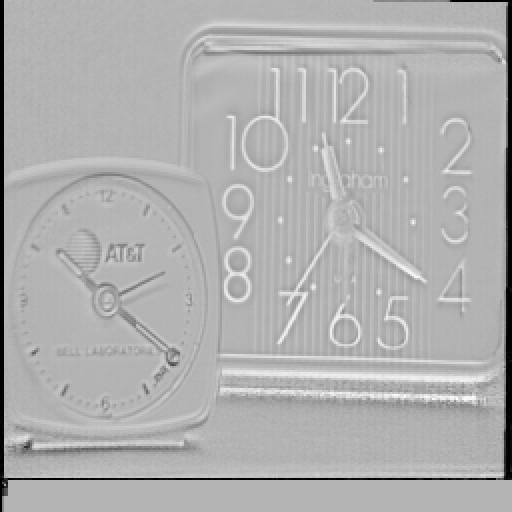} \\
    \end{tabular}
\caption{Source image and sensor gain examples. The first two images are the sources. The third and fourth are their corresponding gains computed by PCA method.}
    \label{fig 1}
\end{figure}

Typically, image fusion is viewed as a problem of generating the fused image from several source images. However, the fusion task can also be posed as an inverse problem. It assumes that the fused image $X_{0}$ exists already, and the image formation model describes how to obtain the source images $X_{i}$ from the fused image as $X_{i} = f(X_{0})$, where $f$ is related to some degradation operations.
When the source images are viewed as signals coming from different sensors, the problem can be modeled as \cite{2010Total}. 
\begin{equation}
    \label{eq1}
    X_{i} = \beta_{i}X_{0} + N_{i}
\end{equation}
\noindent where $\beta_{i}$ and $N_{i}$ are the sensor gain and sensor noise of the $i^{th}$ source image. $X_{0}$ is the fused result. In this paper, the sensor gain $\beta_{i}$ is estimated from the source images $X_i$ employing a patch-based principal component analysis (PCA) strategy. The effectiveness and robustness of PCA on computing sensor gains have been proven over the years~\cite {2010Total, 7025431}. The gains $\beta_{i}$ (see Fig. \ref{fig 1}) constitute the forward operators that transform $X_{0}$ into the source images. The goal, therefore, becomes to minimize the differences between each source image and the degraded fused image, as shown in (\ref{eq2}).
\begin{equation}
    \label{eq2}
    L = \|{X_{1} - \beta_{1}X_{0}}\|_2^2 + \|{X_{2} - \beta_{2}X_{0}}\|_2^2
\end{equation}
\subsection{DIP-based Image Fusion}
DIP is a classical encoder-decoder style network where only one image is directly employed to compute the loss. Therefore, it is always applied in single image processing problems. Although Ulyanov \etal~ \cite{2017Deep} have provided an example of using DIP to reconstruct an image from a pair of flash and no-flash images, their method is not suitable for most image fusion tasks where we want to harvest information from all the source images optimally. In their method, the flash image is applied as the network input, and only the no-flash one is employed to compute the loss. Therefore, their strategy is not straightforwardly extended to multi-image problems. 

\begin{figure}[htb]
\centering
    \includegraphics[width = \columnwidth]{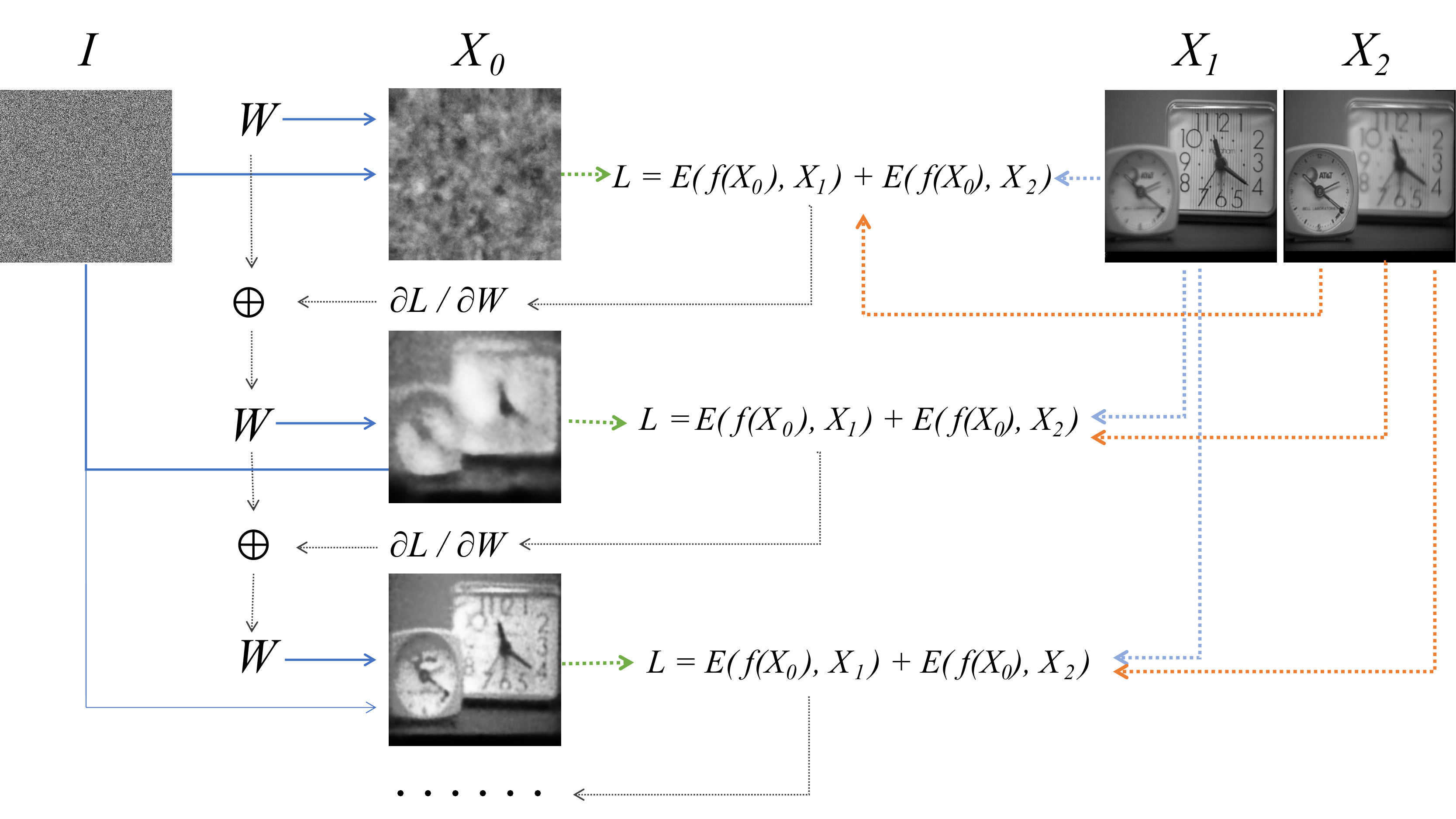}
\caption{The image fusion process. $I$ is the input of the network. $W$ and $X_0$ are the parameter and output of the network, respectively. $X_1$ and $X_2$ are the source images. $E()$ is the loss function corresponding to each input. $f()$ is the degradation function and $L$ is the overall loss.}
\label{fig 2}
\end{figure}

This paper proposes to directly compute the loss from all source images. As shown in Fig. \ref{fig 2}, the input of the network, $I$, is a random tensor that has the same shape as the source images $X_1$ and $X_2$. After passing the input through the DIP architecture (expressed as the weights matrix $W$), the output $X_0$ is generated (\ref{eq3}).
\begin{equation}
    \label{eq3}
    X_{0} = WI
\end{equation}
When formulating the image fusion task as an inverse problem discussed in the previous section, the loss of the network can be computed from (\ref{eq2}).
Subsequently, the backpropagation updates $W$, and a new iteration begins. After a judiciously chosen number of iterations, the output with minimum loss value will be selected as the fused image.

Our loss function offers DIP a chance to harvest information meaningfully from all source images. Furthermore, this structure is easily generalized to fusing images with more than two sources. Additionally, This method trains from scratch for each set of source images and only harvests the underlying features of the current sources. This inherently addresses the lack of training data in image fusion, and therefore overfitting is no longer an issue.
\subsection{Network Architecture}
\label{sec:Network Setting}
This paper employs the super-resolution structure of DIP from \cite{2017Deep} as the foundation of the network. The hyperparameters are kept as their default values. Both the encoder and decoder include 5 convolutional layers of 128 dimensions. For each skip connection, only 4 dimensions are left as the aid. The learning rate is set to 0.01. LeakyReLU and Adam are applied as the activation function and optimizer, respectively. The input and initialization of the network are done randomly. We attempted to perform the task on deeper networks with higher dimensions for each layer, and we also tried different learning rates, activation functions, optimizers. However, they do not help improve the results significantly.
\subsection{Multi-Channel Training Strategy}
The original DIP network only provides a single channel output. In order to capture image priors as exhaustively as possible, we apply a multi-channel strategy to train the network. When the total number of output channels is set to $n$, we copy the source images into $n$ channels. The input becomes a $n$-channel random tensor, and the output is adjusted to $n$ channels as well. The loss function is still calculated by (\ref{eq2}) for each output channel. Subsequently, an averaging strategy is employed to get the final result $X_{0}^{\prime}$, as shown in (\ref{eq4}), where  $X_{0}^{i}$ is the $i^{th}$ channel of the output.
 \begin{equation}
    \label{eq4}
    X_{0}^{\prime} = \frac{\sum_{i=1}^{n}X_{0}^{i}}{n}
\end{equation}
 This alternative is an efficient way to train the network several times and combine the results to achieve better performance. As shown in Fig. \ref{fig 3}, our multi-channel strategy outperforms the single channel method of the original DIP network.
\section{Experiments and Evaluation}
\label{sec:evaluation}
\subsection{Experiments}
\label{sec:Baselines}
The experiments rely on 12 greyscale image pairs of different categories: 1 multi-focus image pair, 1 pair of infrared and visible images, 9 pairs of Magnetic Resonance (MR) and Computed Tomography (CT) from different patients, and a pair of MR and Ultrasound (US) phantom images. The MR and US image pair was obtained through the combination of a beefsteak and a polyvinyl alcohol (PVA) phantom \cite{2020Fusion}. 
The results are compared with three state-of-the-art image fusion methods, one model-based strategy and two deep learning based methods. In the model-based one~\cite{2020Image}, the authors formulated the problem as an inverse one and applied non-convex penalties to get better results than many state-of-the-art methods. The second benchmark \cite{2018Infrared} decomposed source images into base and detail parts (Base and Detail Net). They applied a VGG net trained on ImageNet to harvest deep features for the detail parts and fused the features finally. Our last benchmark is DenseFuse~\cite{2018DenseFuse}. They added a dense block into a pre-trained encoder to capture deep features exhaustively before fusing the features using some strategies. The second benchmark can only process 256 $\times$ 256 images, so we resize all source images. The number of training iterations is set to 2000, an empirical number obtained by trial and error.
\begin{figure}[htb]
\centering
    \includegraphics[width=\columnwidth]{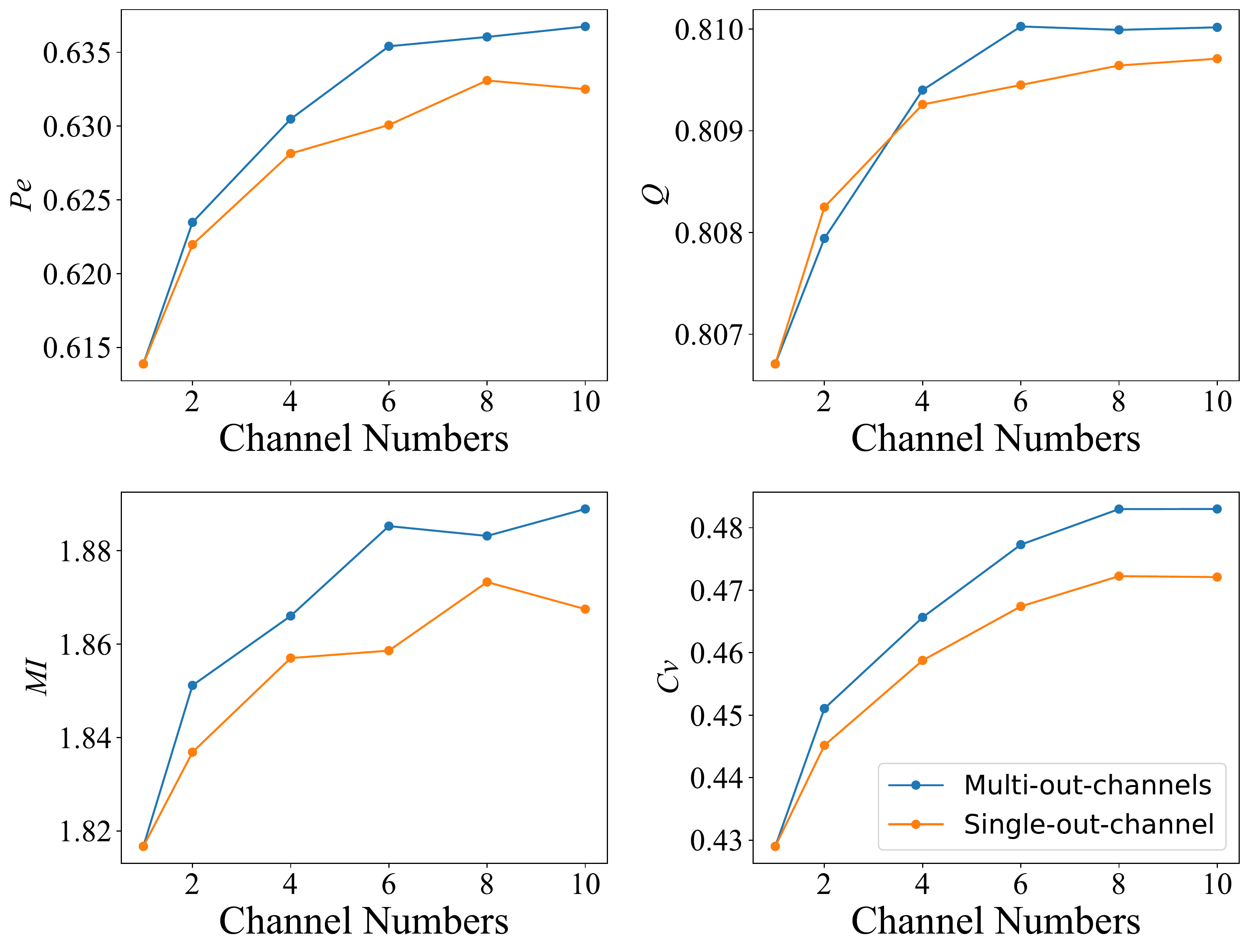}
\caption{The average objective results, on the 12 experiment image pairs, of our multi-channel training strategy and the single-channel method of the original DIP network.}
\label{fig 3}
\end{figure}
\begin{figure*}[htb]
    \centering
    \begin{tabular}{c@{\hspace{0.1pt}}c@{\hspace{5pt}}c@{\hspace{0.1pt}}c@{\hspace{0.1pt}}c@{\hspace{0.1pt}}c}
        \includegraphics[width=0.15\linewidth]{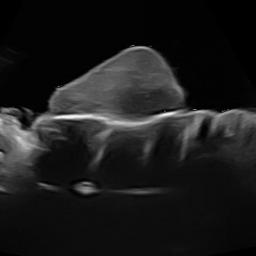}&
        \includegraphics[width=0.15\linewidth]{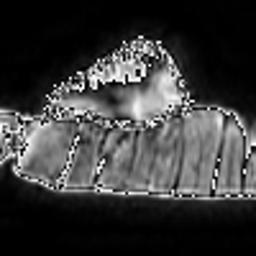}&
        \includegraphics[width=0.15\linewidth]{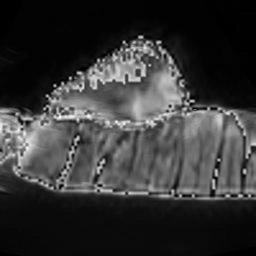}&
        \includegraphics[width=0.15\linewidth]{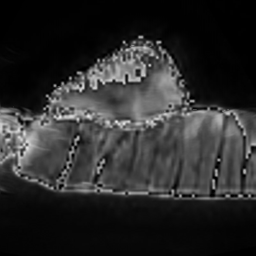}&
        \includegraphics[width=0.15\linewidth]{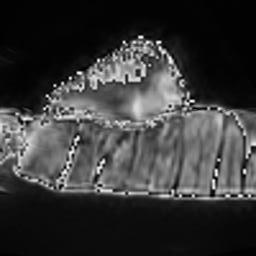}&
        \includegraphics[width=0.15\linewidth]{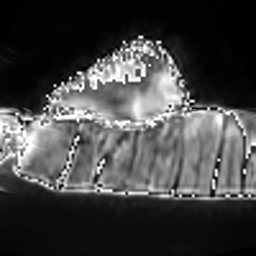} \\
        \includegraphics[width=0.15\linewidth]{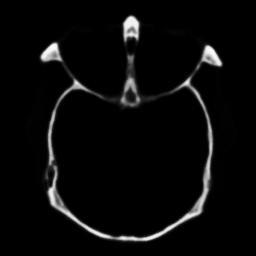}&
        \includegraphics[width=0.15\linewidth]{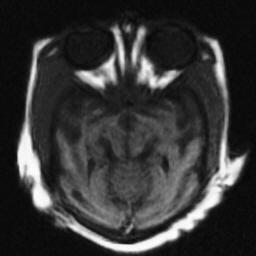}&
        \includegraphics[width=0.15\linewidth]{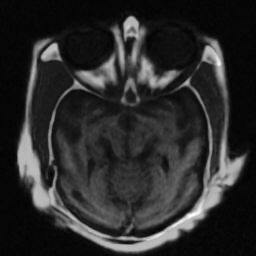}&
        \includegraphics[width=0.15\linewidth]{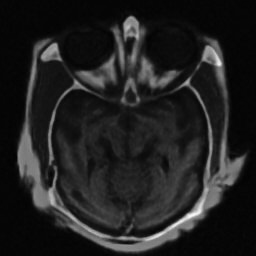}&
        \includegraphics[width=0.15\linewidth]{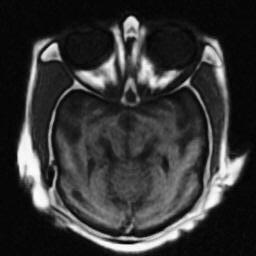}&
        \includegraphics[width=0.15\linewidth]{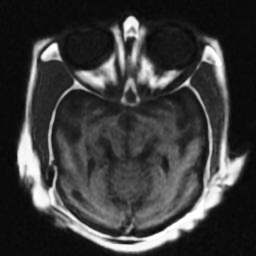} \\
         \includegraphics[width=0.15\linewidth]{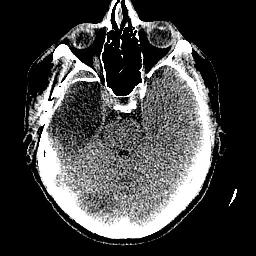}&
        \includegraphics[width=0.15\linewidth]{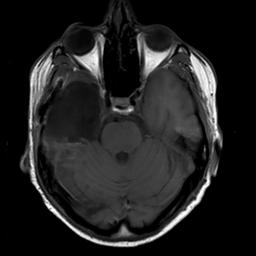}&
        \includegraphics[width=0.15\linewidth]{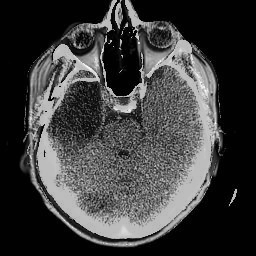}&
        \includegraphics[width=0.15\linewidth]{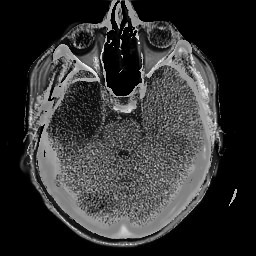}&
        \includegraphics[width=0.15\linewidth]{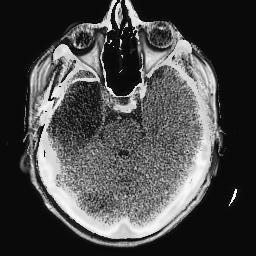}&
        \includegraphics[width=0.15\linewidth]{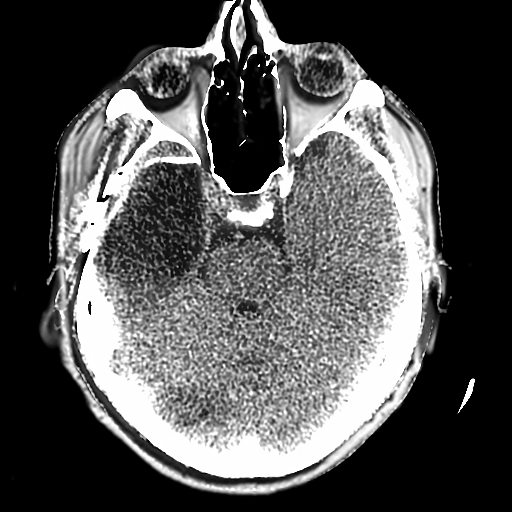} 
    \end{tabular}
\caption{The results of different fusion methods. From left to right, two source images and the results for the non-convex penalty based strategy, the Base and Detail Net, the DenseFuse Net and our 10-channel training DIP strategy, respectively. The Source image in the first row is a pair of MR and US phantom images obtained through the combination of a beefsteak and a polyvinyl alcohol (PVA) phantom. The other two rows are CT and MR image pairs of the human head.}
\label{fig 4}
\end{figure*}
\subsection{Objective and Visual Evaluation}
This paper employs four commonly used numerical measures for image fusion to evaluate the performance objectively: Petrovic and Xydeas's metric ($Pe$) \cite{2000Objective}, Mutual Information ($MI$) \cite{2000Image}, Piella's metric ($Q$) \cite{2003A} and Cvejic's Quality Index ($Cv$) \cite{2005A}. $Pe$ quantifies how much important visual information from the source images is transferred in the fused image. $Q$ is a metric considering local information and the level of distortion. $MI$ focuses on the shared information present in both the fused and the source images. $Cv$ is based on a universal image quality index, and it uses local measurements to quantify the importance of the information. None of them is suitable for all images and can assess the fusion performance comprehensively. Therefore, this paper employs all of them to make the evaluation more comprehensive. 

The results corresponding to training using the different number of channels are shown in Fig. \ref{fig 3}. The results on all image pairs are averaged for each channel number. It is clear that our multi-channel strategy performs better on all metrics than the single channel method of the original DIP. This ascertains that our method can better capture the underlying image statistics than the original DIP. Also, it objectively proves the efficiency of our DIP-based multi-channel training strategy.

\begin{table}[tb]
\caption {The objective results. Average operation is carried out to the CT and MR image pairs to save space. The highest scores are bold and the second-highest scores are shown in blue.} 
\centering
\begin{adjustbox}{width=1\linewidth}
\Large
\begin{tabular}{cccccc}
\hline
%  \multirow{2}{*} &  & \multirow{2}{*}{Non-Convex Penalty} & \multirow{2}{*}{Base and Detail Net} & \multirow{2}{*}{DenseFuse} & \multirow{2}{*}{Our 10-channel Method} \\ \\ \hline 
  &    & \begin{tabular}{c}Non-Convex\\ Penalty \end{tabular} & \begin{tabular}{c} Base and \\Detail Net \end{tabular}& DenseFuse & \begin{tabular}{c} Our 10-channel \\ Method \end{tabular}\\ \hline
\multirow{4}{*}{Multi focus}        & $Pe$ & 0.6297	& \textbf{0.6483}	& 0.6291	& {\color{blue}0.6375}     \\  
                                    & $MI$ & \textbf{3.1107}	& {\color{blue}3.0852}	& 3.0052	& 3.063              \\  
                                    & $Q$  & 0.8843	& \textbf{0.9049}	& {\color{blue}0.8916}	& 0.8487              \\  
                                    & $Cv$ & 0.7278	& \textbf{0.7909}	& {\color{blue}0.7779}	& 0.6991              \\ \hline
\multirow{4}{*}{VIS and IR}         & $Pe$ & {\color{blue}0.4606}     & \textbf{0.4639} & 0.4338  & 0.4464                \\  
                                    & $MI$ & {\color{blue}1.4172 }    & 1.2546    & \textbf{1.4817}      & 1.3017       \\ 
                                    & $Q$  & {\color{blue}0.6932}     & \textbf{0.7279} & 0.6657 & 0.6658              \\  
                                    & $Cv$ & {\color{blue}0.6941}    & \textbf{0.7085}  & 0.6126   & 0.6785               \\ \hline
\multirow{4}{*}{Phantom Data}       & $Pe$ & 0.5468     & 0.4957    & {\color{blue}0.5934 }     & \textbf{0.6625}      \\  
                                    & $MI$ & {\color{blue}1.9511 }   & 1.6075   & 1.8493      & \textbf{1.9533}      \\ 
                                    & $Q$  & 0.8078     & 0.7932    & {\color{blue}0.831}     & \textbf{0.8474}      \\  
                                    & $Cv$ & 0.5555     & {\color{blue}0.6005}   & 0.5476     & \textbf{0.6152}      \\ \hline   
\multirow{4}{*}{CT MR Average}          & $Pe$ & 0.5349      & 0.4383          & {\color{blue}0.5723}   & \textbf{0.6548}   \\  
                                    & $MI$ &  \textbf{1.9769}  & 1.6052  & 1.6962         & {\color{blue}1.8177}   \\ 
                                    & $Q$  & 0.7761  & 0.7263  &{\color{blue} 0.7994}     & \textbf{0.8176}    \\  
                                    & $Cv$ & \textbf{0.4325} & 0.3798 & 0.3664   & {\color{blue}0.4223}   \\ \hline
\end{tabular}
\end{adjustbox}

\label{tab 1} 
\end{table}
We illustrate the performance of different methods in table \ref{tab 1}. The results of 9 pairs of CT and MR images are averaged to save space. It is clear that our 10-channel training strategy achieves the highest or second-highest scores in more than half of the cases. Moreover, Our approach shows apparent advantages compared to others when fusing medical images. For example, we improve the value of $Pe$ by about $15\%$ over the second-best method on both the phantom and real medical images. Especially on the noisy phantom data, our approach shows the best results on all metrics. Although the benchmarks outperform us on several assessments for the other two image pairs, our results are trailing very close behind.

As shown in Fig. \ref{fig 4}, visually, our fused images (the $6^{th}$ column) can capture the bright information from the source images better and show better contrast than the benchmarks. When the source images are extremely noisy, such as the phantom data (the $1^{st}$ row), this strength is more apparent.
\section{Conclusions}
\label{sec:conclusion}
This paper introduces an unsupervised method for image fusion. We pose the task as an inverse problem and enable DIP to be used in multi-sensor image fusion problems. Our method implicitly solves the problem of data scarcity in previous deep learning based methods. Furthermore, as all training is conducted specifically for the source images at hand, overfitting and model mismatch problems in existing deep learning based image fusion methods are eliminated. Additionally, the use of the multi-channel training strategy further strengthens the ability of DIP to capture image statistical priors. As a result, our approach is more robust to noise than its competitors. Visually, we provide fusion results that are clearer and have better contrast than the benchmarking state-of-the-art conventional and deep learning based methods. Objectively, the method proposed in this paper achieves the best performance in terms of several metrics. The advantages are particularly evident when fusing medical images especially when noisy. Finally, Our approach can be easily extended to fuse image datasets containing more than two images, and our multi-channel training strategy is also applicable to other deep learning based problems.

% Below is an example of how to insert images. Delete the ``\vspace'' line,
% uncomment the preceding line ``\centerline...'' and replace ``imageX.ps''
% with a suitable PostScript file name.
% -------------------------------------------------------------------------

%\vfill
%\pagebreak

\bibliographystyle{IEEEbib}
\bibliography{DIP_Fusion}

\end{document}